# Ensemble Methods for Convex Regression with Applications to Geometric Programming Based Circuit Design

**Lauren A. Hannah**  LH140@DUKE.EDU
Duke University, Durham, NC 27708 USA

**David B. Dunson**  DUNSON@STAT.DUKE.EDU
Duke University, Durham, NC 27708 USA

## Abstract

Convex regression is a promising area for bridging statistical estimation and deterministic convex optimization. New piecewise linear convex regression methods (Hannah and Dunson, 2011; Magnani and Boyd, 2009) are fast and scalable, but can have instability when used to approximate constraints or objective functions for optimization. Ensemble methods, like bagging, smearing and random partitioning, can alleviate this problem and maintain the theoretical properties of the underlying estimator. We empirically examine the performance of ensemble methods for prediction and optimization, and then apply them to device modeling and constraint approximation for geometric programming based circuit design.

## 1. Introduction

Convex regression, which is regression subject to a convexity or concavity constraint on the mean function, has received renewed attention. The regression problem is $\mathbf{x} \in \mathcal{X} \subset \mathbb{R}^p$ and $y \in \mathbb{R}$,

$$y = f_0(\mathbf{x}) + \epsilon,$$

where $\epsilon$ is a mean 0 random variable and $f_0$ is convex,

$$\lambda f_0(\mathbf{x}_1) + (1-\lambda) f_0(\mathbf{x}_2) \geq f_0(\lambda \mathbf{x}_1 + (1-\lambda)\mathbf{x}_2),$$

for every $\mathbf{x}_1, \mathbf{x}_2 \in \mathcal{X}$ and $\lambda \in (0,1)$. Regression problems with known convexity or concavity constraints occur in many areas, including economic production, consumption and preference functions (Allon et al., 2007), options pricing (Aït-Sahalia and Duarte, 2003; Hannah and Dunson, 2011), value function approximation in operations research and reinforcement learning (Powell, 2007; Lim, 2010) and device modeling for geometric programming based circuit design in electrical engineering (Kim et al., 2004; Roy et al., 2007).

Convex regression is particularly promising for the machine learning community as a way to bridge statistical estimation and deterministic convex optimization. In particular, data can be used to estimate constraints or objective functions for convex optimization problems. For instance, many reinforcement learning problems that involve resource allocation or storage have concave value functions. If value function estimates are concave, vector-valued continuous action spaces can easily be searched. Similarly, geometric programming and other deterministic convex optimization problems require known constraint and objective functions. However, in many situations only noisy samples are available; convex regression can be used to generate those functions from samples.

Although convex regression has been studied since the 1950's (Hildreth, 1954), computationally feasible methods for the multivariate setting have only recently been proposed by Magnani and Boyd (2009) and Hannah and Dunson (2011). Both methods fit a piecewise linear model to the data, $(\mathbf{x}_1, y_1), \ldots, (\mathbf{x}_n, y_n)$, under a least squares objective function by adaptively partitioning the dataset. While efficient, the method of Magnani and Boyd (2009) does not always converge; the Convex Adaptive Partitioning (CAP) method of Hannah and Dunson (2011), however, converges, is consistent and has a worst case computational complexity of $\mathcal{O}(n \log(n)^2)$.

While piecewise linear methods are computationally efficient, the number of components and hyperplane parameters can be sensitive to training data. Moreover, the resulting piecewise linear models may not be appropriate for approximating functions for convex optimization. When a piecewise linear function is used in an optimization setting, it defines a polyhedral constraint region. When the objective function is linear, a solution lies on a vertex of the polyhedral constraint region. A vertex is created by the intersection of $p+1$ hyperplanes. Because all of the parameters have estimation error, the location of the vertex can be highly sensitive to training data.

These problems can be addressed by using ensemble methods based on the CAP estimator. Ensemble methods combine multiple models to produce a new predictive model. We average over multiple piecewise linear estimators to create a new estimator that is less sensitive to individual hyperplane parameters. Efficient estimators can be created by using many traditional ensemble methods, like bagging (Breiman, 1996), smearing (Breiman, 2000) and random forests (Breiman, 2001), that maintain the properties of the underlying estimator, like consistency and computational complexity. We compare these methods with CAP and the piecewise linear model of Magnani and Boyd (2009); the ensemble methods have better predictive error and produce functions that are much more stable in an optimization setting.

We apply ensemble methods to device and constraint modeling for circuit optimization via geometric programming. Circuits are an interconnection of electrical devices, including capacitors, resistors, inductors and logic gates. Circuit optimization selects appropriate sizes for devices, gates, wires and other design variables, such as threshold and power supply voltage to minimize a given objective like circuit delay or physical area, subject to a set of constraints, usually on area, power, noise or delay. Many circuit design problems can be well modeled by a geometric program (GP), which minimizes a posynomial objective function subject to posynomial inequality and monomial equality constraints. Geometric programming allows the efficient computation of optimal global solutions, even for large problems; see Boyd, Kim, Patil and Horowitz (2005) for a tutorial. However, constraint functions are often not available in a posynomial form and must be approximated from observational data or a known but non-posynomial function for each device. We use ensemble methods to compute device models that are more accurate and more stable than other convex regression methods in this setting.

The contributions of this paper are 1) new ensemble methods for convex regression that are resistant to overfitting and produce a better estimator for optimization than non-ensemble methods, 2) conditions for consistency when CAP is the underlying estimator, 3) strong empirical results, and 4) an application to device and constraint modeling for geometric programming based circuit optimization.

## 2. Ensemble convex regression

### 2.1. Convex regression

Regression subject to a convexity constraint has been the subject of renewed interest in the past few years. One approach has been approximation by a function with a positive definite Hessian (Roy et al., 2007; Aguilera and Morin, 2009; Yongqiao and He, 2012). These methods generally result in a problem that is solved by a semidefinite program with $n$ semidefinite constraints; solution methods are prohibitively slow for more than 1 or 2 thousand observations. Another approach relies on an alternate definition of convexity:

$$f_0(\mathbf{x}_1) \geq f_0(\mathbf{x}_2) + g_0(\mathbf{x}_1)^T(\mathbf{x}_1 - \mathbf{x}_2), \qquad (1)$$

for every $\mathbf{x}_1, \mathbf{x}_2 \in \mathcal{X}$, where $g_0(\mathbf{x}) \in \partial f_0(\mathbf{x})$ is a subgradient of $f_0$ at $\mathbf{x}$. Equation (1) means that a convex function lies above all of its supporting hyperplanes; with enough supporting hyperplanes, $f_0$ can be approximately reconstructed arbitrarily well by taking the maximum over those hyperplanes.

The least squares estimator (LSE) directly projects a least squares objective function onto the cone of convex functions (Hildreth, 1954),

$$\min \sum_{i=1}^{n}(y_i - \hat{y}_i)^2 \quad \text{s.t.} : \hat{y}_j \geq \hat{y}_i + \mathbf{g}_i^T(\mathbf{x}_j - \mathbf{x}_i), \quad (2)$$

for $i, j = 1, \ldots, n$. Here, $\hat{y}_i$ and $\mathbf{g}_i$ are the estimated values of $f_0(\mathbf{x}_i)$ and the subgradient of $f_0$ at $\mathbf{x}_i$, respectively. Equation (2) is a quadratic program with $\mathcal{O}(n^2)$ constraints and cannot be solved efficiently for more than 1 or 2 thousand observations.

To combat these computational difficulties, some recent methods (Magnani and Boyd, 2009; Aguilera et al., 2011; Hannah and Dunson, 2011) estimate a small set of hyperplanes by approximately solving

$$(\alpha^*, \beta^*, K^*) = \arg\min_{\alpha,\beta,K} \sum_{i=1}^{n}\left[y_i - \max_{k=1,\ldots,K} \alpha_k + \beta_k^T \mathbf{x}_i\right]^2$$

where $(\alpha, \beta) \in \mathbb{R} \times \mathbb{R}^p$ defines a hyperplane. The regression function $\hat{f}$ is defined as the maximum over the set of hyperplanes for a convex function,

$$\hat{f}(\mathbf{x}) = \max_{k=1,\ldots,K^*} \alpha_k^* + \beta_k^{*T}\mathbf{x}.$$

The most computationally efficient methods are given by Magnani and Boyd (2009) and Hannah and Dunson (2011).

### 2.2. Combining ensemble methods with convex regression

Ensemble methods reduce overfitting by averaging over a collection of estimates. Here we overview traditional ensemble methods and discuss how they can be combined with convex regression.

**Bagging.** Bagging methods (Breiman, 1996) subsample the training data with replacement, which acts as a random re-weighting of the training set. We study the situation where $M$ new training sets are created by subsampling $n$ observations, denoted by $(\mathbf{x}^{(m)}, \mathbf{y}^{(m)})_{m=1}^M$. An estimator is created by averaging the function estimates for each subsample,

$$\hat{f}^{avg}(\mathbf{x}) = \frac{1}{M} \sum_{m=1}^M \hat{f}\left(\mathbf{x} \,|\, \mathbf{x}^{(m)}, \mathbf{y}^{(m)}\right), \qquad (3)$$

where $\hat{f}\left(\mathbf{x} \,|\, \mathbf{x}^{(m)}, \mathbf{y}^{(m)}\right)$ is a convex regression estimator trained on $(\mathbf{x}^{(m)}, \mathbf{y}^{(m)})$. Bagging can be used with both CAP and the method proposed in Magnani and Boyd (2009).

**Smearing.** Smearing (Breiman, 2000) works by adding mean zero, i.i.d. noise to the training data responses. These "smeared" datasets are then fit with a regression method and the results are averaged to produce an estimator. $M$ new training response sets, $(\mathbf{y}_{1:n}^{(m)})_{m=1}^M$, are created by adding i.i.d. Gaussian noise,

$$\xi^{(m)} \sim N_n(0, \sigma^2 I), \qquad y_i^{(m)} = y_i + \xi_i^{(m)}.$$

The ensemble estimator is then created by averaging convex estimators for each of the $M$ random training sets as in Equation (3). In Breiman (2000), the noise level was chosen to be 2.5 times the standard deviation of the estimator residuals, $\mathbf{y} - \hat{f}(\mathbf{x})$. However, we also consider the situation where it is chosen by cross-validation. Both CAP and the method of Magnani and Boyd (2009) can be used with smearing.

**Random Search Directions.** Random forests (Breiman, 2001) are used in tree regression settings; instead of fully exploring each of the subset split directions, a split is generated in a random direction. Since the subsets in both CAP and Magnani and Boyd (2009) are defined by the hyperplane parameters, there is no direct analogy between random forests and these methods. However, we propose a method in the same spirit.

The partitions of the CAP estimator are created in a two step process. In the first step, subsets are split along cardinal (CAP) or random directions. In the second step, the subsets are redefined by the maximal hyperplanes. Searching over a set of random directions in the first step produces a random estimator. This is done $M$ times and an ensemble estimator is produced by averaging the estimators.

**Boosting.** Boosting (Freund and Schapire, 1997; Friedman, 2002) is a popular ensemble method that constructs additive models in a greedy, forward stepwise manner. This exact method is not appropriate for convex regression since the residuals left after fitting a convex function may not maintain convexity. However, methods that iteratively weight a set of basis functions may prove useful for convex regression.

## 3. Theoretical results

Bagging, smearing and random search directions maintain consistency if CAP is used as the convex estimator and a few mild conditions are imposed. Each CAP covariate subset $A_k$ has diameter $d_{nk}$, where $d_{nk} = \sup_{\mathbf{x}_1, \mathbf{x}_2 \in A_k} \|\mathbf{x}_1 - \mathbf{x}_2\|_2$. Define the empirical mean for index subset $C_k$ as $\bar{x}_k = \frac{1}{|C_k|} \sum_{i \in C_k} \mathbf{x}_i$. For $\mathbf{x}_i \in A_k$, define

$$\Gamma_i^{(m)} = \begin{bmatrix} [1, \ldots, 1] \\ d_{nk}^{-1}(\mathbf{x}_i - \bar{\mathbf{x}}_k) \end{bmatrix}, \quad G_k = \sum_{i \in C_k} \Gamma_i \Gamma_i^T.$$

Let $\mathbf{x}_1, \ldots, \mathbf{x}_n$ be i.i.d. random variables and let a superscript of $(m)$ denote that a quantity is associated with random estimator $m = 1, \ldots, M$. Let $\hat{f}(\mathbf{x} \,|\, Z^{(m)}, D_n)$ be a random estimator based on data $D_n$ and random variable $Z^{(m)}$. We make the following assumptions, which are the original CAP conditions for consistency applied to each random dataset:

**A1.** $\mathcal{X}$ is compact and $f_0$ is Lipschitz continuous and continuously differentiable on $\mathcal{X}$ with Lipschitz parameter $\zeta$.

**A2.** There is an $a > 0$ such that $\mathbb{E}\left[e^{a|Y - f_0(x)|} \,|\, \mathbf{X} = \mathbf{x}\right]$ is bounded on $\mathcal{X}$.

**A3.** For $m = 1, \ldots, M$, the diameter of the partition $\max_k d_{nk}^{(m)^{-1}} \to 0$ in probability as $n \to \infty$.

**A4.** Let $\lambda_k^{(m)}$ be the smallest eigenvalue of $|C_k^{(m)}|^{-1} G_k^{(m)}$ and $\lambda_n^{(m)} = \min_k \lambda_k^{(m)}$. Then for $m = 1, \ldots, M$, $\lambda_n^{(m)}$ remains bounded away from 0 in probability as $n \to \infty$.

**A5.** For $m = 1,\ldots,M$, the number of observations in each subset satisfies $\min_{k=1,\ldots,K_n^{(m)}} d_{nk}^{(m)2} |C_k^{(m)}|/\log(n) \to 0$ in probability as $n \to \infty$.

**Proposition 3.1.** *Suppose that*

$$\hat{f}^{avg}(\mathbf{x} \mid D_n) = \frac{1}{M} \sum_{m=1}^{M} \hat{f}(\mathbf{x} \mid Z^{(m)}, D_n)$$

*and* $\sup_{\mathbf{x} \in \mathcal{X}} \left| \hat{f}(\mathbf{x} \mid Z^{(m)}, D_n) - f_0(\mathbf{x}) \right| \to 0$ *in probability. Then for every fixed $M$, $\sup_{\mathbf{x} \in \mathcal{X}} \left| \hat{f}^{avg}(\mathbf{x} \mid D_n) - f_0(\mathbf{x}) \right| \to 0$ in probability.*

*Proof.* By the triangle inequality,

$$\left| \hat{f}^{avg}(\mathbf{x} \mid D_n) - f_0(\mathbf{x}) \right| \le \frac{1}{M} \sum_{m=1}^{M} \left| \hat{f}(\mathbf{x} \mid Z^{(m)}, D_n) - f_0(\mathbf{x}) \right|,$$

The result follows from the assumption for each $m$. □

**Theorem 3.2.** *If $\hat{f}(\mathbf{x})$ is generated by the CAP estimator, $(\mathbf{x}^{(m)}, \mathbf{y}^{(m)})_{m=1}^{M}$ are generated by bagging, $\mathbb{E}[Y^2 \mid X = x] < \infty$ a.s. for all $\mathbf{x} \in \mathcal{X}$, and **A1.** to **A5.** hold, then for every fixed $M$,*

$$\sup_{\mathbf{x} \in \mathcal{X}} \left| \frac{1}{M} \sum_{m=1}^{M} \hat{f}(\mathbf{x} \mid \mathbf{x}_{1:n}^{(m)}, \mathbf{y}_{1:n}^{(m)}) - f_0(\mathbf{x}) \right| \to 0$$

*in probability as $n \to \infty$.*

*Proof.* Because of the bounded second moment and **A1.**, the plugin estimate $\hat{f}(\mathbf{x} \mid \mathbf{x}^{(m)}, \mathbf{y}^{(m)})$ is consistent and the result follows from Prop. 3.1. □

**Theorem 3.3.** *If $\hat{f}(\mathbf{x})$ is generated by the CAP estimator, $Y_i^{(m)} = Y_i + \xi_i^{(m)}$ with $\xi_i^{(m)} \sim N(0, B)$ iid for some $B < \infty$, and assumptions **A1.** through **A5.** hold, then for every fixed $M$,*

$$\sup_{\mathbf{x} \in \mathcal{X}} \left| \frac{1}{M} \sum_{m=1}^{M} \hat{f}(\mathbf{x} \mid \mathbf{x}_{1:n}, \mathbf{y}_{1:n}^{(m)}) - f_0(\mathbf{x}) \right| \to 0$$

*in probability as $n \to \infty$.*

*Proof.* Fix $m$ and consider the estimator $\hat{f}(\mathbf{x} \mid \mathbf{x}_{1:n}, \mathbf{y}_{1:n}^{(m)})$. Note that $Y^{(m)} = Y + \xi^{(m)}$, so

$$\mathbb{E}\left[e^{a|Y^{(m)} - f_0(x)|} \mid \mathbf{x}\right] \le \mathbb{E}\left[e^{a(|Y - f_0(x)| + |\xi^{(m)}|)} \mid \mathbf{x}\right]$$

$$\le 2 e^{\frac{1}{2}a^2 B^2} \mathbb{E}\left[e^{a|Y - f_0(x)|} \mid \mathbf{x}\right] < \infty.$$

Since **A1.** through **A5.** hold for that estimator, it is consistent and the result follows from Prop. 3.1. □

**Theorem 3.4.** *If $\hat{f}(\mathbf{x})$ is generated by the CAP estimator with random search directions $Z^{(m)}$ and **A1.** to **A5.** hold, then for every fixed $M$,*

$$\sup_{\mathbf{x} \in \mathcal{X}} \left| \frac{1}{M} \sum_{m=1}^{M} \hat{f}(\mathbf{x} \mid \mathbf{x}_{1:n}, \mathbf{y}_{1:n}, Z^{(m)}) - f_0(\mathbf{x}) \right| \to 0$$

*in probability as $n \to \infty$.*

*Proof.* Each estimator is consistent and the result follows from Prop. 3.1. □

## 4. Experiments on synthetic data

### 4.1. Prediction

Here $\mathbf{x} \in \mathbb{R}^5$ with $\mathbf{X} \sim N_5(0, I)$. Set

$$y = (x_1 + .5 x_2 + x_3)^2 - x_4 + .25 x_5^2 + \epsilon, \quad \epsilon \sim N(0, 1).$$

We compared CAP, linear fitting (Magnani and Boyd, 2009) (MB), cross-validated smearing (Sm CAP, Sm MB), smearing with 2.5 times residual noise (Sm 2.5 CAP, Sm 2.5 MB), bagging (Bag CAP, Bag MB) and random search directions (RD). All ensemble methods except RD were implemented with CAP and MB. 10 training sets and one testing set were generated; the number of training samples was varied between 100 and 5,000. For Sm CAP and Sm MB, the noise level was chosen by 5-fold cross validation from $\sigma = \{0, 10^{-2}s, 10^{-1}s, 5^{-1}s, 2.5^{-1}s, s, 2.5s, 5s, 10s, 10^2 s\}$, where $s$ is the standard deviation of the residuals; each level was approximated with $M = 25$. Appropriate noise levels were then probabilistically chosen for each $m$ for $M = 200$. The number of hyperplanes in linear fitting was chosen through 5-fold cross validation. Results are given in Table 2.

Ensemble methods substantially reduced CAP prediction error for all sample sizes except for $n = 5,000$. Smearing with cross-validated noise, random search directions and bagging produced similar results. Ensemble methods produced smaller reduction in prediction error for linear fitting, with bagging producing the best results. Smearing with 2.5x standard deviation noise produced worse results than the other ensemble methods, likely because the noise levels for cross-validated smearing were lower.

### 4.2. Optimization

Approximating objective functions or constraints for use in convex optimization is one of the most promising applications for convex regression. In this subsection, we use convex regression for response surface methods in stochastic optimization; see Lim (2010) for an overview. We would like to minimize an unknown function $f(\mathbf{x})$ with respect to $\mathbf{x}$ given $n$ noisy observations, $(\mathbf{x}_i, y_i)_{i=1}^n$, where $y_i = f(\mathbf{x}_i, \epsilon_i)$,

$$\min_{\mathbf{x} \in \mathcal{X}} \mathbb{E}\left\{ f(\mathbf{x}, \epsilon) \,|\, (\mathbf{x}_i, y_i)_{i=1}^n \right\}. \quad (4)$$

| Method | n | RMSE | Solution Val |
|---|---|---|---|
| CAP | 100 | $0.205 \pm 0.043$ | $0.089 \pm 0.012$ |
| MB | 100 | $0.649 \pm 0.518$ | $0.207 \pm 0.034$ |
| LSE | 100 | $11.333 \pm 9.504$ | $0.077 \pm 0.010$ |
| Sm CAP | 100 | $0.196 \pm 0.039$ | $0.075 \pm 0.009$ |
| Sm MB | 100 | $0.213 \pm 0.118$ | $0.094 \pm 0.021$ |
| Sm 2.5 CAP | 100 | $0.168 \pm 0.034$ | $\mathbf{0.022 \pm 0.003}$ |
| Sm 2.5 MB | 100 | $0.178 \pm 0.039$ | $0.028 \pm 0.004$ |
| Bag CAP | 100 | $0.159 \pm 0.033$ | $0.035 \pm 0.004$ |
| Bag MB | 100 | $\mathbf{0.131 \pm 0.027}$ | $0.041 \pm 0.004$ |
| RD | 100 | $0.175 \pm 0.034$ | $0.052 \pm 0.006$ |
| CAP | 500 | $0.106 \pm 0.014$ | $0.057 \pm 0.007$ |
| MB | 500 | $0.235 \pm 0.174$ | $0.086 \pm 0.015$ |
| LSE | 500 | $1.656 \pm 1.213$ | $0.037 \pm 0.005$ |
| Sm CAP | 500 | $0.100 \pm 0.014$ | $0.046 \pm 0.006$ |
| Sm MB | 500 | $0.106 \pm 0.048$ | $0.044 \pm 0.005$ |
| Sm 2.5 CAP | 500 | $\mathbf{0.071 \pm 0.013}$ | $\mathbf{0.007 \pm 0.001}$ |
| Sm 2.5 MB | 500 | $0.135 \pm 0.071$ | $0.009 \pm 0.002$ |
| Bag CAP | 500 | $0.076 \pm 0.013$ | $0.019 \pm 0.003$ |
| Bag MB | 500 | $0.079 \pm 0.017$ | $0.020 \pm 0.003$ |
| RD | 500 | $0.087 \pm 0.013$ | $0.034 \pm 0.004$ |

Table 1. Root mean squared error (RMSE) and values of approximated solutions plus/minus one standard error for Equation (4).

We tested the regression methods with

$$Y_i = \mathbf{x}_i \mathbf{Q} \mathbf{x}_i^T + \epsilon_i, \quad \mathbf{Q} = \begin{bmatrix} 1 & 0.2 \\ 0.2 & 1 \end{bmatrix}, \quad \epsilon_i \sim N(0, 0.1).$$

The constraint set is $-1 \leq x_j \leq 1$ for $j = 1, 2$, and $\mathbf{x}_i \sim Unif[-1, 1]^2$. We used the above methods as well as the Least Squares Estimator (LSE). We generated 50 training sets, solved Equation (4) using each regression method, and then calculated the root mean squared error (RMSE) for the functional estimators; the number of training data was set at 100 and 500. Solutions to Equation (4) were evaluated with respect to the true function; RMSE was calculated over a grid on the constraint space. Results are given in Table 1. The ensemble methods produced significantly better quality results both in terms of solution selection and RMSE than the competing methods. The extra noise of 2.5x smearing acted as a smoother and produced a more accurate and stable minimum.

## 5. Circuit design

### 5.1. Geometric programming and circuit design

Geometric programming is a mathematical optimization problem where the objective function and constraints are defined in terms of monomials, posynomials and generalized posynomials. A monomial $g(\mathbf{x})$ and posynomial $f(\mathbf{x})$ have the forms

$$g(\mathbf{x}) = c x_1^{a_1} x_2^{a_2} \ldots x_p^{a_p}, \quad f(\mathbf{x}) = \sum_{k=1}^K c_k x_1^{a_{k1}} \ldots x_p^{a_{kp}},$$

for $\mathbf{x} > 0$. A generalized posynomial is created through positive powers, addition, multiplication or the maximum of posynomials. A GP minimizes a generalized posynomial subject to a set of generalized posynomial inequality and monomial equality constraints,

$$\min f_0(\mathbf{x}), \quad \text{subject to } f_i(\mathbf{x}) \leq 1, \quad g_j(\mathbf{x}) = 1, \quad \mathbf{x} > 0,$$

where $f_i$ are generalized posynomials for $i = 0, \ldots, m$ and $g_j$ are monomials for $j = 1, \ldots, k$. GPs can be reformulated as convex optimization problems through a change of variables, $z_i = \log(x_i)$,

$$f(\mathbf{x}) = \sum_{k=1}^K c_k x_1^{a_{k1}} \ldots x_p^{a_{kp}}, \quad f(\mathbf{z}) = \sum_{k=1}^K e^{\mathbf{a}_k^T \mathbf{z} + b_k}.$$

If we take the log of the transformed function, $f(\mathbf{z})$, we get a function that is convex in $\mathbf{z}$.

Many circuit design problems, both analog and digital, can be modeled as GPs (Kim et al., 2004; Boyd, Kim, Patil and Horowitz, 2005; Roy et al., 2007). Geometric programming offers a fast, global solution method for design problems that scales well even to large problems. To use geometric programming, however, devices and constraints need to be modeled by generalized posynomials. Sometimes device models are not known and need to be inferred from data in standard cell libraries; other times, constraints or device models are given, but not in a form that can be expressed as a generalized posynomial. In each of these cases, piecewise linear convex regression can be used to produce generalized posynomial representations of these models and constraints. Ensemble methods can produce models that have lower error and are more stable in an optimization setting than existing methods.

### 5.2. Device and constraint modeling with convex regression

Functions without explicit generalized posynomial representation occur in two settings. In the device modeling setting, device parameters such as the inverse of transconductance, gate-source voltage, the inverse of output resistance and the intrinsic gate capacitance need to be modeled as generalized posynomial functions of input parameters such as device width, length, terminal voltages and drain current. These relationships need to be inferred from data generated by circuit simulation or contained in standard cell libraries.

| Method | Prob | $n = 200$ | $n = 500$ | $n = 1,000$ | $n = 2,000$ | $n = 5,000$ |
|---|---|---|---|---|---|---|
| CAP | Syn | $1.027 \pm 0.218$ | $0.703 \pm 0.121$ | $0.408 \pm 0.050$ | $0.332 \pm 0.054$ | $0.195 \pm 0.026$ |
| MB | Syn | $0.699 \pm 0.123$ | $0.444 \pm 0.071$ | $0.360 \pm 0.062$ | $0.269 \pm 0.047$ | $0.178 \pm 0.029$ |
| Sm CAP | Syn | $0.800 \pm 0.106$ | $0.421 \pm 0.067$ | $0.328 \pm 0.039$ | $0.269 \pm 0.035$ | $0.210 \pm 0.022$ |
| Sm MB | Syn | $0.622 \pm 0.110$ | $0.391 \pm 0.063$ | $0.312 \pm 0.048$ | $0.234 \pm 0.045$ | $0.154 \pm 0.023$ |
| Sm 2.5 CAP | Syn | $0.897 \pm 0.109$ | $0.517 \pm 0.087$ | $0.352 \pm 0.041$ | $0.292 \pm 0.029$ | $0.236 \pm 0.021$ |
| Sm 2.5 MB | Syn | $0.807 \pm 0.124$ | $0.456 \pm 0.071$ | $0.333 \pm 0.047$ | $0.264 \pm 0.041$ | $0.184 \pm 0.036$ |
| Bag CAP | Syn | $0.809 \pm 0.113$ | $0.432 \pm 0.072$ | $0.334 \pm 0.038$ | $0.279 \pm 0.032$ | $0.212 \pm 0.023$ |
| Bag MB | Syn | $\mathbf{0.603 \pm 0.091}$ | $\mathbf{0.369 \pm 0.058}$ | $\mathbf{0.282 \pm 0.036}$ | $\mathbf{0.216 \pm 0.035}$ | $\mathbf{0.142 \pm 0.016}$ |
| RD | Syn | $0.807 \pm 0.107$ | $0.421 \pm 0.064$ | $0.325 \pm 0.039$ | $0.269 \pm 0.035$ | $0.211 \pm 0.021$ |
| CAP | Pow | $0.144 \pm 0.031$ | $0.125 \pm 0.012$ | $0.056 \pm 0.010$ | $0.049 \pm 0.010$ | $0.024 \pm 0.003$ |
| MB | Pow | $0.040 \pm 0.014$ | $0.019 \pm 0.005$ | $0.014 \pm 0.003$ | $0.010 \pm 0.002$ | $0.009 \pm 0.002$ |
| Sm CAP | Pow | $0.131 \pm 0.025$ | $0.090 \pm 0.013$ | $0.052 \pm 0.008$ | $0.030 \pm 0.004$ | $0.019 \pm 0.003$ |
| Sm MB | Pow | $\mathbf{0.038 \pm 0.01}$ | $\mathbf{0.015 \pm 0.003}$ | $\mathbf{0.012 \pm 0.002}$ | $\mathbf{0.009 \pm 0.001}$ | $\mathbf{0.007 \pm 0.001}$ |
| Sm 2.5 CAP | Pow | $0.135 \pm 0.025$ | $0.091 \pm 0.014$ | $0.051 \pm 0.008$ | $0.030 \pm 0.004$ | $0.019 \pm 0.002$ |
| Sm 2.5 MB | Pow | $0.061 \pm 0.019$ | $0.049 \pm 0.022$ | $0.026 \pm 0.010$ | $0.022 \pm 0.009$ | $0.029 \pm 0.015$ |
| Bag CAP | Pow | $0.134 \pm 0.025$ | $0.093 \pm 0.013$ | $0.053 \pm 0.008$ | $0.030 \pm 0.004$ | $0.020 \pm 0.002$ |
| Bag MB | Pow | $0.042 \pm 0.015$ | $0.017 \pm 0.003$ | $0.014 \pm 0.003$ | $0.010 \pm 0.002$ | $\mathbf{0.007 \pm 0.001}$ |
| RD | Pow | $0.131 \pm 0.025$ | $0.092 \pm 0.014$ | $0.051 \pm 0.027$ | $0.030 \pm 0.012$ | $0.020 \pm 0.007$ |

*Table 2.* Root mean squared error (RMSE) plus/minus one standard error for the synthetic prediction problem in Section 4.1 (Syn) and power modeling in Section 5.3 (Pow) by approximation method and training sample size.

Log convexity of such data is not guaranteed and the goal is to find a low error generalized posynomial approximation. In the constraint modeling setting, some constraints, such as the power supply voltage for different gates, are known but do not have a generalized posynomial form. In this case, the goal is to create a low error generalized posynomial approximation based on samples from the true function. While posynomials can be directly fit through Gauss-Newton type methods, these methods only reach local optima and are sensitive to algorithm initialization. Piecewise linear convex regression offers a more appealing alternative.

Monomials, posynomials and generalized posynomials are closely related to affine (linear) and convex functions. Using the transformation, $\mathbf{z} = \log(\mathbf{x})$, if a function $f$ is a monomial, then $\log(f(e^{\mathbf{z}}))$ is affine. If a function $f$ is a generalized posynomial, then $\log(f(e^{\mathbf{z}}))$ is convex. Conversely, if $\log(f(e^{\mathbf{z}}))$ is convex, then $f$ can be approximated arbitrarily well by a posynomial, generalized posynomial or the maximum of a set of monomials. We use this fact and piecewise linear convex regression to approximate what should be posynomial functions by the maximum of a set of monomials.

Let $(\mathbf{x}_i, y_i)_{i=1}^n$ be a set of observations of $f(\mathbf{x})$ and let $(\mathbf{z}_i, \hat{y}_i)_{i=1}^n$ be the set of transformed observations, $\mathbf{z}_i = \log(\mathbf{x}_i)$ and $\hat{y}_i = \log(y_i)$. If a piecewise linear convex model is fit to the transformed data,

$$\hat{f}_n(\mathbf{z}) = \max_{k=1,\ldots,K} \alpha_k + \beta_k^T \mathbf{z},$$

then a generalized posynomial can be constructed for the original function,

$$\tilde{f}_n(\mathbf{x}) = \max_{k=1,\ldots,K} e^{\alpha_k} x_1^{\beta_{k1}} \ldots x_p^{\beta_{kp}}.$$

In an ensemble setting, the coefficients can either be constructed directly or with dummy variables.

### 5.3. Power modeling

Here we fit a generalized posynomial to a known, but non-posynomial, function, the power dissipated as a function of gate supply and threshold voltages, $V_{dd}$ and $V_{th}$; this example is studied in Boyd, Kim, Patil and Horowitz (2005). The total power dissipated for a gate is the sum of the average static power and the dynamic power dissipated. The dynamic power is a function of the gate supply voltage, $V_{dd}$,

$$P_{dyn} = f\left(C^{int} + C^L\right) V_{dd}^2,$$

where $f$ is the frequency, $C^{int}$ is the intrinsic capacitance and $C^L$ is the load capacitance. The static leakage is a function of the supply voltage and the average current leakage, $\bar{I}^{leak}$, $P_{stat} = \bar{I}^{leak} V_{dd}$. The average current leakage is a function of both the supply and threshold voltages; a standard model is

$$\bar{I}^{leak} \propto e^{-(V_{th} - \gamma_D V_{dd})/V_0},$$

where $\gamma_D$ and $V_0$ are constants, typically around 0.06 and 0.04, respectively. To get the total power dissipates, we set $P = P_{stat} + P_{dyn}$. Note that while $P_{dyn}$ is a posynomial, $P_{stat}$ is not; moreover, it is not even convex under the log transformation. Previous methods have modeled power dissipated through hand-tuned monomial and generalized posynomial approximations. As a numerical example, we would like to model

$$P = V_{dd}^2 + 30 V_{dd} e^{-(V_{th} - 0.06 V_{dd})/0.039} \tag{5}$$

for $1.0 \leq V_{dd} \leq 2.0$ and $0.2 \leq V_{th} \leq 0.4$ with a generalized posynomial. The goal is to produce a model that has low overall error.

We produce a generalized posynomial model using $n$ covariate samples that are drawn uniformly from $x_1 \in [\log(1.0), \log(2.0)]$ and $x_2 \in [\log(0.2), \log(0.4)]$; responses are generated by evaluating those values in Equation (5). The number of observations was varied between 200 and 5,000, with 10 i.i.d. training sets. Methods were the same as in Section 4.1. All were tested by calculating RMSE from Equation (5) on a 10,000 sample testing set. For the ensemble methods, $M = 200$. Results are given in Table 2. The gains using ensemble methods were smaller for power modeling than for the synthetic problem, likely because the power modeling problem is noiseless.

### 5.4. LC oscillator design

Here we compare convex regression methods for device modeling in geometric programming based LC oscillator design. Oscillators generate an oscillating output at a constant frequency. An LC oscillator (L and C represent inductor and capacitor, respectively) sends electrons from one plate of a capacitor through a coil, or loop inductor, to reach the other plate. However, when the electrons travel around a coil, a magnetic field is created that generates a voltage across the coil in the opposite direction of the electron flow. Once the capacitor is fully discharged, the magnetic field around the coil collapses and the voltage recharges the capacitor in the opposite direction. Additional voltage is applied to compensate for that lost to resistance.

We implement the LC oscillator design problem given in Boyd, Kim and Mohan (2005). The goal is to minimize power consumption subject to upper bound constraints on the phase noise, area of the loop inductor and lower bounds on the loop gain and the self resonance frequency, and some other loop inductor and transistor-specific constraints. The variables to be optimized are the width and diameter of the loop inductor, the self resonance frequency, the length and width and maximum current of the CMOS transistor, the differential voltage amplitude, the total capacitance of the oscillator, the maximum switching capacitance, the minimum variable capacitance, the bias current and the capacitor max frequency.

We used the methods in Section 4.1 to approximate the resistance of the loop inductor. EM-based posynomial fitting gave

$$R = 0.1 DW^{-1} + 3 \cdot 10^{-6} DW^{-0.84} f^{0.5} \qquad (6)$$
$$+ 5 \cdot 10^{-9} DW^{-0.76} f^{0.75} + 0.02 DW f,$$

where $f$, $D$, $W$ are the loop inductor frequency, diameter and width. We used this model as truth to compare the suboptimality of the regression methods in a non-trivial optimization setting. To generate approximations, we sampled uniformly across log-transformed covariates, where $-10 \leq \log(D) \leq -5$, $-13 \leq \log(W) \leq -10$ and $22 \leq \log(f) \leq 23$ for $n = 500$ and $n = 5,000$, although $f$ is fixed for optimization. 50 different training sets were generated and the optimization problem was solved using the approximated functions in `ggplab` (Mutapcic et al., 2006). For all ensemble methods, $M = 50$ to limit the number of non-sparse GP constraints. We compared optimal power consumption for each model as a function of phase noise (dBc/Hz), which was varied from $-122$ to $-110$. Due to differences in function scale, percentage error from optimal values and percentage deviations from the true resistance values are used instead of RMSE. Results are given in Table 3.

For both methods, cross-validated smearing provided lower error estimators and produced solutions with comparable mean deviation and lower maximum deviation. Random search directions produced similar results with CAP. The different measurement metric highlights the differences between ensemble methods. The Magnani and Boyd (2009) estimator became more unstable with 2.5x standard deviation smearing. CAP became more unstable when used with bagging. Both of these methods likely do poorly because they add significant noise into a noiseless situation.

## 6. Conclusions

In this paper, we combine convex regression (CAP and Magnani and Boyd (2009)) and ensemble methods to produce a piecewise linear approximation method. CAP is a consistent, stable estimator that uses a tree-like search to produce a piecewise linear model. Ensemble methods like bagging, smearing and random search directions add uncertainty in the partition boundaries. When averaged, these uncertain estimates produce a better fit. The Magnani and Boyd (2009) method is an unstable estimator that can produce a very good fit by aligning a piecewise linear model with the data used to produce it. Smearing and bagging average over a large number of models and reduce the likelihood that the estimator will be determined by a few poorly fitting models. Although device modeling is a natural setting for ensemble convex regression, the low computational complexity, theoretical guarantees and strong empirical performance in optimization settings, make ensemble convex regression a promising tool for combining estimation and optimization.

Ensemble Methods for Convex Regression

| Method | Mean Dev | Max Dev | Mean Sol | Max Sol | Mean Dev | Max Dev | Mean Sol | Max Sol |
|---|---|---|---|---|---|---|---|---|
| | | $n = 500$ | | | | $n = 5,000$ | | |
| CAP | $2.39 \pm 0.01$ | $13.29 \pm 0.21$ | $5.93 \pm 0.15$ | $7.86 \pm 0.21$ | $0.44 \pm 0.02$ | $3.68 \pm 0.22$ | $0.90 \pm 0.07$ | $2.95 \pm 0.19$ |
| MB | $6.66 \pm 0.95$ | $35.53 \pm 2.74$ | **$0.60 \pm 0.04$** | $1.83 \pm 0.10$ | $2.32 \pm 0.32$ | $21.67 \pm 2.36$ | **$0.70 \pm 0.03$** | $1.49 \pm 0.05$ |
| Sm CAP | $1.41 \pm 0.02$ | $12.13 \pm 0.22$ | $1.12 \pm 0.07$ | $6.50 \pm 0.18$ | $0.23 \pm 0.00$ | $3.68 \pm 0.07$ | $0.97 \pm 0.02$ | $1.39 \pm 0.04$ |
| Sm MB | $0.18 \pm 0.02$ | $3.61 \pm 0.90$ | $0.73 \pm 0.04$ | $1.66 \pm 0.07$ | **$0.07 \pm 0.00$** | **$1.26 \pm 0.05$** | $0.86 \pm 0.02$ | **$1.38 \pm 0.03$** |
| Sm 2.5 CAP | $1.51 \pm 0.02$ | $12.57 \pm 0.23$ | $1.23 \pm 0.08$ | $7.07 \pm 0.20$ | $0.24 \pm 0.00$ | $3.82 \pm 0.06$ | $0.96 \pm 0.02$ | $1.48 \pm 0.05$ |
| Sm 2.5 MB | $2.49 \pm 1.85$ | $20.69 \pm 14.23$ | $1.44 \pm 0.85$ | $2.77 \pm 1.05$ | $3.14 \pm 2.96$ | $19.89 \pm 17.74$ | $6.53 \pm 5.56$ | $17.29 \pm 15.68$ |
| Bag CAP | $1.42 \pm 0.02$ | $12.37 \pm 0.21$ | $1.16 \pm 0.08$ | $6.61 \pm 0.16$ | $1.41 \pm 0.02$ | $12.37 \pm 0.22$ | $1.16 \pm 0.08$ | $6.75 \pm 0.17$ |
| Bag MB | **$0.13 \pm 0.00$** | **$3.26 \pm 0.10$** | $0.87 \pm 0.02$ | **$1.44 \pm 0.02$** | $0.13 \pm 0.00$ | $3.21 \pm 0.12$ | $0.89 \pm 0.02$ | $1.41 \pm 0.02$ |
| RD | $1.40 \pm 0.02$ | $12.23 \pm 0.22$ | $1.15 \pm 0.06$ | $6.52 \pm 0.17$ | $0.24 \pm 0.00$ | $3.72 \pm 0.06$ | $0.96 \pm 0.02$ | $1.45 \pm 0.04$ |

*Table 3.* Mean and maximum percentage deviation from true resistance function (Mean Dev, Max Dev) and mean and maximum percentage deviation from solution value using true resistance function (Mean Sol, Max Sol) plus/minus one standard error as a function of approximation method and training sample size.

## Acknowledgments

Lauren A. Hannah is partially supported by the Duke Provost's Postdoctoral Fellowship. This work was supported by Award Number R01ES17240 from the National Institute of Environmental Health Sciences. The content is solely the responsibility of the authors and does not necessarily represent the official views of the National Institute of Environmental Health Sciences or the National Institutes of Health.